\documentclass[10pt,twocolumn,letterpaper]{article}

\usepackage{wacv}              

\definecolor{wacvblue}{rgb}{0.21,0.49,0.74}
\usepackage{algorithm}
\usepackage{algorithmic}

\usepackage{graphicx}
\usepackage[accsupp]{axessibility}

\usepackage{amsmath}
\usepackage{amssymb}
\usepackage{booktabs}
\usepackage{times}
\usepackage{epsfig}
\usepackage{amssymb}
\usepackage{enumitem}  
\usepackage{hyperref}
\usepackage{multirow}
\usepackage{multicol}
\usepackage{siunitx}
\usepackage{tabularx}
\usepackage[table]{xcolor} 

\definecolor{softblue}{HTML}{EAF4FB} 

\usepackage{array}
\usepackage{comment}
\usepackage{bm}
\usepackage{tikz}
\usepackage{makecell}
\usepackage{pifont}
\usepackage{fancyhdr}
\usepackage{arydshln}
\usepackage{verbatim}
\usepackage{pifont}
\usepackage{footmisc}
\usepackage{float}
\usepackage{algorithm, algorithmic}
\definecolor{lightblue}{rgb}{0.6, 0.6, 0.8}
\usepackage{microtype}
\def\wacvPaperID{2513} 
\def\confName{WACV}
\def\confYear{2027}

\title{Video Generalized Category Discovery}

\author{Bardia Safaei\\
Johns Hopkins University\\
{\tt\small bsafaei1@jhu.edu}
\and
Vishal M. Patel\\
Johns Hopkins University\\
{\tt\small vpatel36@jhu.edu}
}
\begin{document}
\maketitle
\thispagestyle{empty}
\pagestyle{empty}
Generalized Category Discovery (GCD) aims to categorize unlabeled data from both \textit{known} and \textit{unknown} classes using labeled data from only the known classes. While recent GCD methods have shown promise on static images, their performance degrades in videos due to temporal dynamics, larger intra-class variation, and the lack of temporal consistency in existing approaches. We introduce \textbf{VideoGCD}, a framework tailored for video action recognition under the GCD setting. Our method integrates (i) a partitioned Mixture-of-Experts module that encourages specialization for known and novel categories, (ii) a novel-class balancing loss to reduce bias toward known classes, and (iii) a temporal optimal transport loss that promotes consistent routing across frames. Experiments on three video benchmarks demonstrate that VideoGCD significantly improves performance over existing GCD methods, particularly on novel categories. Our findings highlight the importance of combining temporal modeling and expert specialization for video GCD.    
\section{Introduction}
In recent years, deep learning has achieved remarkable success in various visual recognition tasks, such as classification \cite{lu2007survey, safaei2023open}, segmentation \cite{cao2022swin, yu2023techniques}, and object detection \cite{he2017mask, vs2023towards}, by training models on large datasets with full supervision. However, this approach assumes a closed-world setting where the categories seen during training match those encountered at test time. In many real-world applications such as autonomous driving or video surveillance, this assumption rarely holds. New classes may appear during deployment, and manual annotation is often costly or impractical. Generalized Category Discovery (GCD) \cite{vaze2022generalized} addresses this challenge by leveraging a small labeled dataset of \textit{known} classes together with a large unlabeled dataset containing both known and \textit{unknown} classes. The goal is to correctly classify the known categories while discovering and grouping the unknown ones, enabling recognition systems that can adapt without full supervision.

\begin{figure}[t!]
    \begin{center}
        \includegraphics[width=0.90\linewidth]{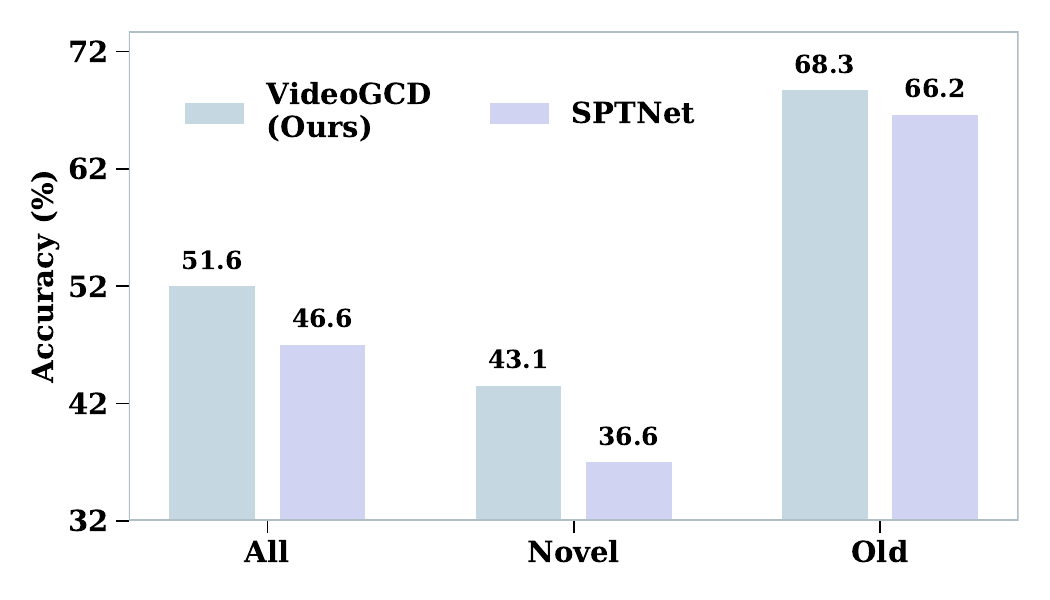}
    \end{center}
    \vskip -15.0pt
    \caption{
    Average clustering accuracy on three benchmark datasets (UCF-101, HMDB-51, and Something-Something-V2) compared to the state-of-the-art method SPTNet. VideoGCD achieves significant improvements in overall performance, particularly on the Novel subset, demonstrating its strong generalization ability for the GCD task.
    }
    \label{fig:motivation} 
    \vskip -8.0pt
\end{figure}

The GCD problem has been explored through contrastive representation learning \cite{vaze2022generalized, pu2023dynamic, zhang2023promptcal, choi2024contrastive}, clustering-based discovery \cite{otholt2024guided, vaze2022generalized}, prototypical parametric classification \cite{wen2023parametric, hu2025prototypical, an2023generalized}, and prompt learning \cite{qu2024adaptgcd, zhang2025less}. While these methods have shown strong performance on static image datasets, their extension to video remains underexplored. Videos introduce unique challenges such as temporal redundancy, stronger intra-class variation, motion blur, and background clutter, all of which amplify pseudo-label noise and make novel category discovery more difficult. Moreover, existing image-based GCD methods do not enforce temporal consistency across frames, which is crucial for coherent representation learning in videos \cite{yan2023temporally}. Compounding these challenges, the GCD setting introduces an additional difficulty in videos: the strong supervision available for labeled known classes often dominates the learning process, causing the model to bias its representations toward known categories while under-utilizing unlabeled samples that may belong to novel classes. In addition, video data exhibits strong temporal redundancy, where frames belonging to the same action should ideally produce consistent representations. However, frame-level features may fluctuate due to background frames, motion blur, or noisy predictions, leading to unstable similarity relationships across time. Such instability can negatively affect clustering and category discovery in video models.

To address these challenges, we propose \textbf{VideoGCD}, a framework for generalized category discovery in video action recognition. VideoGCD introduces a \emph{partitioned mixture-of-experts} architecture that encourages balanced capacity allocation for both known and novel categories while maintaining a shared backbone representation. To stabilize expert utilization during training, we introduce a balancing objective and a novel-class balancing loss, denoted as $\mathcal{L}_{\text{bal}}$ and $\mathcal{L}_{\text{ncb}}$, which encourage equitable expert activation and ensure that unlabeled samples receive sufficient learning signal. To further improve stability in video representations, we introduce a temporal consistency objective based on optimal transport, denoted as $\mathcal{L}_{\text{tot}}$, which aligns routing distributions across consecutive frames and encourages temporally coherent expert assignments. Combined with the standard contrastive learning objective, these components form a unified framework that promotes temporally stable representations for video-based category discovery.


In summary, this work makes the following contributions:
\begin{itemize}
    \item We provide the first systematic analysis of generalized category discovery in the context of video action recognition, highlighting key challenges related to temporal dynamics and feature domain gap.
    \item We propose \textbf{VideoGCD}, a framework that integrates a partitioned Mixture-of-Experts architecture, a novel-class balancing loss, and a temporal optimal transport loss to address key challenges in video GCD.
    \item Through extensive experiments on UCF-101, HMDB-51, and Something-Something-V2, we demonstrate that VideoGCD consistently outperforms state-of-the-art GCD baselines, especially in novel class discovery (see Fig.~\ref{fig:motivation}).
\end{itemize}
\section{Related Work}

\textbf{Generalized Category Discovery (GCD)} GCD \cite{liu2025hyperbolic, vaze2022generalized, yang2024learning, zheng2024textual, tan2024revisiting, tu2024memory} addresses the challenge of learning from partially labeled datasets where the unlabeled set may contain both known and novel categories. Early work, such as ORCA \cite{cao2022openworld} addressed the bias toward seen classes by introducing an uncertainty-adaptive margin that balances the learning of discriminative features for known and novel categories, hence reducing the gap in intra-class variance. DCCL \cite{pu2023dynamic} improved clustering by alternating between estimating visual concepts and learning their representations, using a dynamic concept generation and update mechanism to maintain consistency during training. More recent approaches have moved toward parametric classifiers and improved fine-tuning strategies. In SimGCD \cite{wen2023parametric}, the authors utilize a parametric classification objective with entropy regularization. AdaptGCD \cite{qu2024adaptgcd} extended this idea by introducing adapter tuning for better adaptation to the target dataset. 
Unlike AdaptGCD, which is designed for static images, VideoGCD addresses
video-specific challenges through novel-class expert balancing and temporally
consistent frame-level routing using partition-aware optimal transport. Prompt-based methods such as SPTNet \cite{wang2024sptnet} employed spatial prompts to maintain discriminative features, while SelEx \cite{rastegar2024selex} adopted hierarchical semi-supervised clustering to improve discovery on fine-grained datasets.
While these methods have shown strong performance on static image datasets, they do not explicitly address the additional challenges introduced by video data, such as temporal consistency and larger domain gap, motivating our proposed framework.

\textbf{Video Action Recognition} A wide range of approaches have been explored for video action recognition \cite{wu2017recent, pham2022video, wang2024taylor, wang2023videomae}, each emphasizing different aspects of spatiotemporal modeling. Early convolutional methods include I3D \cite{carreira2017quo} that uses 3D convolutions to capture motion, and TSM \cite{lin2019tsm} which adapts 2D convolutions with temporal shifts. SlowFast \cite{feichtenhofer2019slowfast} employs two parallel pathways: a slow branch that samples frames sparsely to focus on spatial details, and a fast branch that processes frames more densely to better capture motion and other short-term temporal patterns. TPN \cite{yang2020temporal} focused on improving the integration of multi-level features. More recently, transformer-based models such as Video Swin Transformer \cite{liu2022video} and VideoMAE \cite{tong2022videomae} have achieved strong results by leveraging large-scale pretraining to build richer video representations.

\textbf{Mixture of Experts (MoE)}
The Mixture of Experts framework increases model capacity by splitting the input space into multiple regions and assigning each region to a dedicated expert network. A gating module selects the most relevant experts for each input, and their outputs are combined to produce the final prediction \cite{cai2025survey, zhou2022mixture, masoudnia2014mixture}. This design allows experts to specialize on different parts of the data distribution and has been widely adopted in various applications. 

In our framework, we leverage a partitioned MoE architecture to encourage balanced capacity allocation for both known and novel categories in the generalized category discovery setting. The backbone video encoder is adapted using parameter-efficient fine-tuning (LoRA) to allow lightweight adaptation while keeping most pretrained parameters fixed.
\section{Preliminaries}
\label{sec:prelim}
\medskip\noindent\textbf{Video Generalized Category Discovery}. We define a video dataset $\mathcal{D} = \mathcal{D}_L \cup \mathcal{D}_U$, where 
$\mathcal{D}_L = \{(v_i, y_i) \mid v_i \in \mathcal{V}, y_i \in \mathcal{Y}_L\}$ is the labeled subset 
containing videos from a set of known action classes $\mathcal{Y}_L$, and 
$\mathcal{D}_U = \{ v_j \mid v_j \in \mathcal{V} \}$ is the unlabeled subset. 
The labels for $\mathcal{D}_U$ are drawn from 
$\mathcal{Y}_U = \mathcal{Y}_L \cup \mathcal{Y}_N$, where 
$\mathcal{Y}_N = \mathcal{Y}_U \setminus \mathcal{Y}_L$ represents novel classes not seen in 
$\mathcal{D}_L$. We denote the total number of classes as 
$K = |\mathcal{Y}_U| = |\mathcal{Y}_L| + |\mathcal{Y}_N|$. Each video 
$v \in \mathcal{V}$ is a sequence of $T$ frames 
$v = \{ f_1, f_2, \dots, f_T \}$. The goal of Video GCD is to train a model 
$F_\theta(\cdot)$ using both $\mathcal{D}_L$ and $\mathcal{D}_U$ so that it can 
(1) correctly classify videos from the known classes $\mathcal{Y}_L$, and 
(2) automatically discover and cluster videos from the novel classes 
$\mathcal{Y}_N$. This problem is more challenging than the image-based GCD 
setting because videos introduce challenges such as temporal dynamics, long-range dependencies, and intra-class 
variation. Furthermore, 
video backbones are often pretrained on smaller, action-centric datasets (e.g., Kinetics), 
which creates a larger domain gap. These factors make clustering and category discovery significantly harder.

\medskip\noindent\textbf{Overview of SimGCD}. Our method builds on SimGCD \cite{wen2023parametric}, a parametric GCD approach that jointly learns discriminative representations and class prototypes using both labeled and unlabeled data.

\textit{Representation learning.} SimGCD employs supervised and self-supervised contrastive learning. Given two augmented views $v_i$ and $v_i'$ of the same video, the unsupervised contrastive loss is calcualted as:
\[
\mathcal{L}_{\text{unsup}} = \frac{1}{|B|} \sum_{i \in B}
- \log \frac{\exp(z_i^\top z_i' / \tau_u)}{\sum_{j \neq i} \exp(z_i^\top z_j' / \tau_u)} ,
\]
where $z_i = G(F(v_i))$ is the normalized feature from the backbone $F(\cdot)$ and projector $G(\cdot)$, and $\tau_u$ is the temperature. For labeled videos, positives are defined by class labels:
\[
\mathcal{L}_{\text{sup}} = \frac{1}{|B_l|} \sum_{i \in B_l}
\frac{1}{|\mathcal{N}_i|}
\sum_{k \in \mathcal{N}_i}
- \log \frac{\exp(z_i^\top z_k' / \tau_c)}
{\sum_{j \neq i} \exp(z_i^\top z_j' / \tau_c)} ,
\]

where $\mathcal{N}_i$ is the set of labeled videos of the same class. The total representation loss is 
$\mathcal{L}_{\text{rep}} = \lambda \mathcal{L}_{\text{sup}} + (1-\lambda) \mathcal{L}_{\text{unsup}}$.

\textit{Prototypical classification.} A set of prototypes $C = \{c_1, \dots, c_K\}$ represent all known and novel classes. Predictions are obtained via cosine similarity:
\begin{equation}
    p_i^{(k)} = \frac{\exp(h_i^\top c_k / \tau_s)}
{\sum_{k'} \exp(h_i^\top c_{k'} / \tau_s)} ,
\end{equation}

where $h_i = F(v_i)$ is the feature for video $v_i$ and $\tau_s$ is the temperature. The classification loss is 
\[
\mathcal{L}_{\text{cls}} =
\frac{1}{|B_l|} \sum_{i \in B_l} \ell(y_i, p_i)
+ \frac{1}{|B|} \sum_{i \in B} \ell(q_i', p_i)
- \epsilon H(\bar{p}) ,
\]
where $q_i'$ is the soft pseudo-label for unlabeled videos, and $H(\bar{p}) = - \sum_k \bar{p}^{(k)} \log \bar{p}^{(k)}$ is an entropy regularizer on the average class prediction $\bar{p}$ to avoid prototype collapse.

The final SimGCD loss for videos is 
$\mathcal{L}_{\text{base}} = \mathcal{L}_{\text{rep}} + \mathcal{L}_{\text{cls}}$.

\medskip\noindent\textbf{Optimal Transport (OT)}.
OT provides a way to transform one probability distribution into another at the lowest possible cost. 
Given two discrete probability distributions $p \in \mathbb{R}^n$ and $q \in \mathbb{R}^m$, and a cost function $c(x_i, y_j)$ that measures the cost of moving mass from $x_i$ to $y_j$, the OT problem is defined as:
\[
\begin{aligned}
W(p, q) = \min_{\mu \in \mathbb{R}^{n \times m}} & \sum_{i=1}^n \sum_{j=1}^m c(x_i, y_j) \mu(i,j) \\
\text{s.t.} \quad & \sum_j \mu(i,j) = p_i, \quad \sum_i \mu(i,j) = q_j,
\end{aligned}
\]
where $\mu(i,j)$, known as the transport plan, indicates how much mass is moved from $x_i$ to $y_j$. 
The Wasserstein distance $W(p, q)$ gives the total cost of aligning $p$ and $q$. To make the optimization stable and computationally efficient, entropic regularization \cite{clason2021entropic} is usually included, and the problem is solved using the Sinkhorn algorithm \cite{distances2013lightspeed}.
In this work, we use OT along the temporal dimension to match expert assignments across different frames of the same video. 
We construct a partition-aware cost matrix that penalizes inconsistencies both across time and between expert partitions (e.g., known vs.\ novel experts). 
This temporal OT encourages stable and consistent routing over time and reduces noise in the learned video representations.
\begin{figure*}
    \centering
\includegraphics[width=0.80\textwidth]{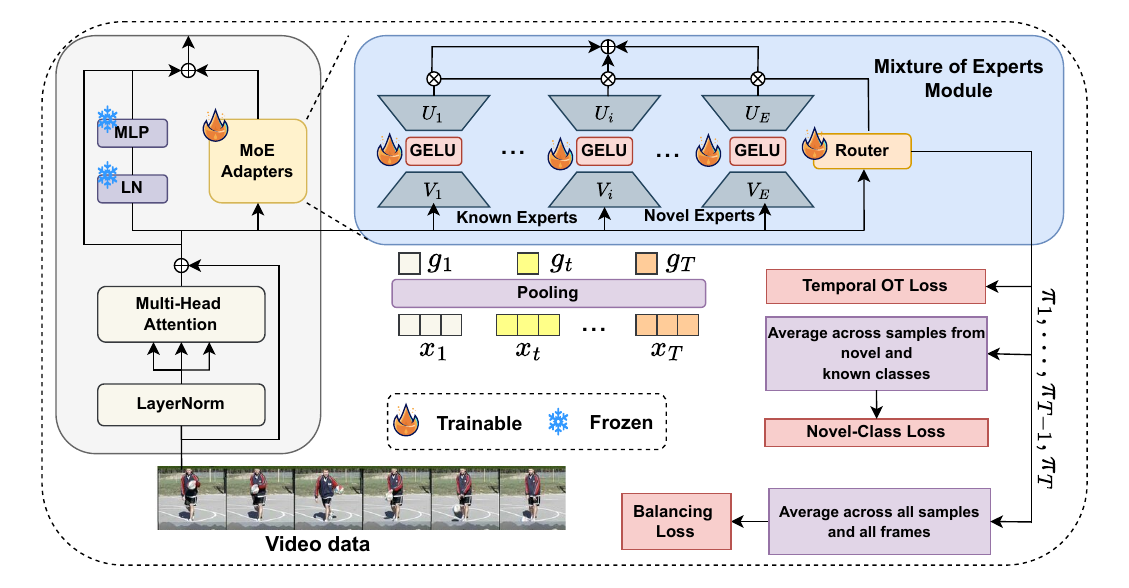}  
\vspace{-5mm}
    \caption{Overview of the proposed \textbf{VideoGCD} framework for Generalized Category Discovery (GCD) in video action recognition. 
Frame-global representations ($g_1, g_t, g_T$) are obtained via temporal pooling and routed through the MoE module, which consists of known and novel experts. 
The router predicts expert assignment probabilities, and each expert applies low-rank adapters ($U_e, V_e$) for parameter-efficient fine-tuning. 
Three auxiliary losses are used: (i) \textit{Balancing Loss} promotes uniform expert usage, (ii) \textit{Novel-Class Balancing Loss} ensures sufficient learning signal for novel-class experts, and (iii) \textit{Temporal OT Loss} enforces consistent expert assignments across frames. The updated representations are integrated back into the transformer backbone. For simplicity, LoRA fine-tuning on the multi-head attention modules is not shown.}
\vspace{-6mm}
    \label{fig:framework} 
\end{figure*}
\section{Proposed Method: VideoGCD}
In this section, we detail our framework for the Video GCD problem. Our approach builds on SimGCD \cite{wen2023parametric}, which has shown strong results on image-based GCD. However, SimGCD is not designed for videos, and in its original form, only finetunes the last transformer block and the projector. While this strategy works for images, it is suboptimal for videos. Video models often face a larger domain gap between pretraining and target datasets and must handle temporal dependencies that SimGCD does not address.
To overcome these issues, we extend the video backbone with MoE modules inserted in the last $M$ layers, inspired by their success in open-world learning frameworks \cite{ du2025mixture, qu2024adaptgcd, yu2024boosting}. We further utilize two balancing losses that stabilize the routing mechanism and ensure that unlabeled novel-class samples receive sufficient training signal, as these samples are typically weaker in supervision. We also introduce a \textit{temporal optimal transport (TOT) loss} to enforce consistent expert assignments across frames. In our implementation, we adapt the pretrained backbone using parameter-efficient fine-tuning (LoRA), while keeping most backbone parameters frozen. (see Fig. \ref{fig:framework}).

\subsection{Mixture-of-Experts (MoE) Design}
We incorporate MoE modules into the last $M$ transformer blocks of the video backbone. Each MoE module operates at the \textit{frame level}, dynamically routing tokens across a set of $E$ experts based on frame-global features as follows.
The same set of $E$ experts is shared across all temporal positions; only the
routing probabilities are computed at the frame level. Given an input sequence of patch tokens $x \in \mathbb{R}^{B \times N \times D}$, where $B$ is the batch size, $N = T \cdot H_p \cdot W_p$ is the total number of tokens, and $D$ is the feature dimension, we reshape $x$ into 
\[
x \in \mathbb{R}^{B \times T \times S \times D}, \quad S = H_p \cdot W_p,
\]
where $T$ is the number of temporal bins and $S$ is the number of spatial patches per frame. For each frame $t = 1, \dots, T$, we compute the \textit{frame-global feature} by averaging over the spatial patches: $g_t = \frac{1}{S} \sum_{s=1}^{S} x_{t,s}$ where $g_t \in \mathbb{R}^{B \times D}.$
Then, a \textit{gating network} $W_g$ predicts the routing weights over $E$ experts for each frame as follows:
\[
\pi_t = \text{softmax}\left( \frac{W_g g_t}{\tau} \right), \quad \pi_t \in \mathbb{R}^{B \times E},
\]
where $W_g \in \mathbb{R}^{E \times D}$ and $\tau$ is a temperature.
Each expert $e$ applies a \textit{low-rank residual transformation} to the frame-global feature as $\Delta_t^{(e)} = U^{(e)} \, \text{GELU} \left( V^{(e)} g_t \right)$
where $V^{(e)} \in \mathbb{R}^{D \times r}$ and $U^{(e)} \in \mathbb{R}^{r \times D}$ are the expert projection matrices with rank $r \ll D$. The value of $r$ is set to 4 in our experiments.
The MoE residual update is a mixture of all experts weighted by their routing probabilities: $\Delta_t = \sum_{e=1}^{E} \pi_t^{(e)} \, \Delta_t^{(e)}.$ Finally, the residual update is broadcast back to all spatial patches of the frame $t$ and added to the input tokens:
\[
y_{t,s} = x_{t,s} + \alpha \, \Delta_t, \quad s = 1, \dots, S.
\]
Here $\alpha$ is a fixed scaling value which we set to 0.1.
This frame-wise design allows each expert to specialize in different parts of the data distribution, while the gating mechanism selects appropriate experts for each frame. 

\subsection{Balancing Loss 
(\texorpdfstring{$\mathcal{L}_{\text{bal}}$})}
The goal of $\mathcal{L}_{\text{bal}}$ is to prevent expert collapse by encouraging a balanced use of all experts, a common objective in the MoE literature \cite{luo2024moelora}. We use the routing probabilities $\pi_t \in \mathbb{R}^{B \times E}$, and compute the average routing probability across the entire batch and all frames as:
\[
\omega_e = \frac{1}{B \cdot T} \sum_{b=1}^{B} \sum_{t=1}^{T} \pi_{b,t}^{(e)}, \quad e = 1, \dots, E.
\]
We then enforce that this average distribution is close to the uniform distribution $q_e = \frac{1}{E}$ where $e = 1, \dots, E$. The loss is defined as the KL divergence between the average routing distribution and the uniform distribution:
\[
\mathcal{L}_{\text{bal}} = D_{\mathrm{KL}}(\omega \; \| \; q)
= \sum_{e=1}^{E} \omega_e \log \frac{\omega_e}{q_e}.
\]
This loss ensures that, on average, all experts are utilized evenly, enhancing the robustness of learned representations. However, it does not explicitly account for the imbalance between labeled known samples and unlabeled samples that may correspond to novel classes. The novel-class balancing loss complements this objective by encouraging separate groups of experts to receive sufficient training signal from both known and novel samples.
$\mathcal{L}_{\mathrm{bal}}$ balances expert utilization rather than the semantic
class distribution. Its statistics are aggregated over $B\times T$ routing
observations and only $E=8$ experts, so full class coverage within each
mini-batch is not required.

\subsection{Novel-Class Balancing Loss 
(\texorpdfstring{$\mathcal{L}_{\text{ncb}}$})}
Since samples from known classes are partially labeled, they receive much stronger supervision than unlabeled novel-class samples, which only benefit from weaker unsupervised signals. As a result, the model can easily become biased toward known classes, diminishing the contribution of novel-class samples to the learned feature representations. To address this issue, we introduce the \textit{novel-class balancing loss} $\mathcal{L}_{\text{ncb}}$, which ensures that experts dedicated to novel classes receive sufficient training signal. We explicitly partition the $E$ experts into two equal groups: the set of known-class experts $\mathcal{E}_{\text{known}}$ and the set of novel-class experts $\mathcal{E}_{\text{novel}}$. For labeled data, we use the ground-truth labels, and for unlabeled data, we use pseudo-labels to determine whether a sample belongs to a known or novel class. We then compute the average routing probabilities for each group:
\[
\begin{aligned}
\omega^{\text{known}}_e &= 
\frac{1}{|\mathcal{D}_{\text{known}}|} 
\sum_{(b,t) \in \mathcal{D}_{\text{known}}} 
\pi_{b,t}^{(e)}, \\
\omega^{\text{novel}}_e &= 
\frac{1}{|\mathcal{D}_{\text{novel}}|} 
\sum_{(b,t) \in \mathcal{D}_{\text{novel}}} 
\pi_{b,t}^{(e)}.
\end{aligned}
\]
We then encourage these distributions to match uniform targets over their respective expert groups:
\[
\begin{aligned}
q^{\text{known}}_e &=
\begin{cases}
\frac{1}{|\mathcal{E}_{\text{known}}|}, & e \in \mathcal{E}_{\text{known}}, \\
0, & \text{otherwise},
\end{cases}
\\[6pt]
q^{\text{novel}}_e &=
\begin{cases}
\frac{1}{|\mathcal{E}_{\text{novel}}|}, & e \in \mathcal{E}_{\text{novel}}, \\
0, & \text{otherwise}.
\end{cases}
\end{aligned}
\]

In practice, the routing probabilities are considered within their corresponding
known or novel expert partition when computing the balancing objective. The loss is finally defined as:
\[
\mathcal{L}_{\text{ncb}} =
D_{\mathrm{KL}}\big(\omega^{\text{known}} \; \| \; q^{\text{known}}\big)
+ D_{\mathrm{KL}}\big(\omega^{\text{novel}} \; \| \; q^{\text{novel}}\big).
\]
This encourages known-class samples to activate the known-class experts and novel-class samples to activate the novel-class experts, ensuring that novel categories have a distinct and reliable learning path.

\subsection{Temporal Optimal Transport Loss (\texorpdfstring{$\mathcal{L}_{\text{tot}}$})}
To enforce temporal consistency in expert routing, we introduce the \textit{Temporal Optimal Transport loss} $\mathcal{L}_{\text{tot}}$. 
This loss encourages consecutive frames from the same video to activate similar experts, thereby stabilizing the learned representations over time. 
Let $\pi_{b,t} \in \mathbb{R}^E$ denote the routing probability vector over the $E$ experts for the $b$-th video in the batch at temporal step $t$. 
For consecutive time steps $t$ and $t+1$, we measure the similarity of their routing distributions using the optimal transport (OT) cost:
\[
\mathcal{C}(p,q) = \min_{\mu \in \Gamma(p,q)} 
\sum_{i=1}^{E} \sum_{j=1}^{E} \mu(i,j) \, C(i,j),
\]
where $p = \pi_{b,t}$ and $q = \pi_{b,t+1}$. 
Here, $\mu \in \mathbb{R}^{E \times E}$ is the \textit{transport plan} that determines how much probability mass is moved from expert $i$ at time $t$ to expert $j$ at time $t+1$.

The set $\Gamma(p,q)$ denotes all valid transport plans that satisfy the marginal constraints $\sum_j \mu(i,j) = p_i$ and $\sum_i \mu(i,j) = q_j$. 
The cost matrix $C \in \mathbb{R}^{E \times E}$ penalizes changes in expert assignments as follows: 
(i) $C(i,i) = 0$ for staying with the same expert, 
(ii) $C(i,j) = c_{\text{in}}$ for switching to another expert within the same partition (known or novel), and 
(iii) $C(i,j) = c_{\text{cross}}$ for switching across partitions. We set $c_{\text{in}}=0.1$ and $c_{\text{cross}}=1$.
This partition-aware structure encourages consistent routing while allowing smooth transitions when necessary.
Finally, the temporal OT loss is averaged across all consecutive temporal bins and all videos in the batch:
\[
\mathcal{L}_{tot} = \frac{1}{B(T-1)} 
\sum_{b=1}^{B} \sum_{t=1}^{T-1} 
\mathcal{C}(\pi_{b,t}, \pi_{b,t+1}).
\]
Minimizing $\mathcal{L}_{tot}$ encourages the routing probabilities of consecutive frames to remain similar, i.e., to have a low OT cost, which improves temporal consistency and the overall feature quality.
\subsection{Overall Loss and Training}
Our final training objective combines the SimGCD loss $\mathcal{L}_{\text{base}}$ with the three MoE losses as follows:
\[
\mathcal{L}_{\text{overall}} = \mathcal{L}_{\text{base}} 
+ \beta \left( \mathcal{L}_{bal} + \mathcal{L}_{ncb} + \mathcal{L}_{tot} \right).
\]
During training, we apply parameter-efficient fine-tuning (LoRA) to the $Q$, $K$, and $V$ projection matrices of the transformer backbone.
This approach allows the model to adapt more effectively to the video GCD setting while preserving the generalization capability 
of the pretrained features.
\section{Experiments}
We conduct extensive experiments on three widely used video action recognition datasets, namely UCF-101 \cite{soomro2012ucf101}, HMDB-51 \cite{kuehne2011hmdb}, and Something-Something-V2 (SS-V2) \cite{goyal2017something}, to demonstrate the effectiveness of VideoGCD. All experiments are run with three different random seeds, and we report the average results. To ensure fair comparisons, we use the exact same pretrained VideoMAE backbone as the initialization for all GCD methods.
\begin{table*}[t]
\centering
\renewcommand{\arraystretch}{1.2}
\caption{Comparison of different GCD methods on SS-V2, UCF-101 and HMDB-51. We report clustering accuracy (\%) on the All, Old, and New class splits. Best results are shown in bold, and the second-best results are underlined. We run each experiment three times and report the average results. All compared methods start from the same pretrained video backbone.}
\vspace{-1mm}
\resizebox{0.70 \linewidth}{!}{
\begin{tabular}{
    l
    >{\columncolor{softblue}}c c c
    >{\columncolor{softblue}}c c c
    >{\columncolor{softblue}}c c c
}
\toprule
\textbf{Method} & \multicolumn{3}{c}{\textbf{SS-V2}} & \multicolumn{3}{c}{\textbf{UCF-101}} & \multicolumn{3}{c}{\textbf{HMDB-51}}  \\
\cmidrule(lr){2-4} \cmidrule(lr){5-7} \cmidrule(lr){8-10}
 & All & Old & New & All & Old & New & All & Old & New \\
\midrule
\textit{$k$-means}~\cite{arthur2006k}& 9.8 & 10.4 & 9.5 & 24.4 & 21.6 & 25.7 & 15.8 & 12.5 & 17.6  \\
RankStats+~\cite{han2021autonovel} & 6.4 & 5.6 & 6.7 & 21.7 & 18.3 & 23.3 & 15.3 & `12.8 & 16.6   \\
GCD~\cite{vaze2022generalized} & 15.4 & 20.9 & 12.8 & 41.6 & 51.7 & 36.8 & 26.3 & 32.3 & 23.0   \\
ORCA~\cite{cao2022openworld} & 25.1 & 35.0 & 20.2 & 61.1 & 77.6 & \underline{53.3} & 39.1 & 57.7 & 29.1   \\
SimGCD~\cite{wen2023parametric} & 31.6 & 40.8 & 27.1 & \underline{64.1} & \underline{87.9} & 52.7 & 42.6 & \underline{70.6} & 27.4   \\

AdaptGCD~\cite{qu2024adaptgcd}& 31.4 & 39.5 & \underline{27.4} & 58.6 & 76.1 & 50.2 & 39.6 & 64.8 & 26.0   \\
SPTNet~\cite{wang2024sptnet} & \underline{31.9} & \underline{41.2} & 27.3 & 61.4 & 87.7 & 48.8 & \underline{44.4} & 67.8 & \underline{31.8}  \\
\rowcolor{softblue}
\textbf{VideoGCD (Ours)} & \textbf{35.5} & \textbf{43.3} & \textbf{31.8} & \textbf{74.5} & \textbf{92.6 }& \textbf{65.8} & \textbf{46.6} & \textbf{71.2} & \textbf{33.4} \\
\bottomrule
\end{tabular}}

\label{main_results}
\end{table*}

\begin{table*}[ht]
\centering
\renewcommand{\arraystretch}{1.2}
\setlength{\tabcolsep}{5pt}
\caption{Ablation study on UCF-101 and HMDB-51 datasets. $\checkmark$ indicates the inclusion of each loss component.}
\begin{tabular}{
    c l c c c c
    >{\columncolor{softblue}}c c c
    >{\columncolor{softblue}}c c c
}
\toprule
\textbf{Index} & \textbf{Method} & LoRA 
& $\mathcal{L}_{bal}$ & $\mathcal{L}_{ncb}$ & $\mathcal{L}_{tot}$
& \multicolumn{3}{c}{\textbf{UCF-101}} 
& \multicolumn{3}{c}{\textbf{HMDB-51}} \\
\cmidrule(lr){7-9} \cmidrule(lr){10-12}
& & & & & & All & Old & New & All & Old & New \\
\midrule
(1) & SimGCD & -  & -- & -- & -- & 64.1 & 87.9 & 52.7 & 42.6 & 70.6 & 27.4 \\
(2) & SimGCD & \checkmark  &  & -- & -- & 69.9 & 88.2 & 61.2 & 43.3 & 66.1 & 30.9 \\
(3) & VideoGCD & \checkmark & \checkmark & -- & -- & 72.6 & 92.4 & 63.1 & 44.5 & 67.2 & 31.7 \\
(4) & VideoGCD & \checkmark & \checkmark & \checkmark & -- & 72.7 & 91.0 & 64.0 & 45.5 & 71.7 & 31.8 \\
\rowcolor{softblue}
(5) & VideoGCD & \checkmark & \checkmark & \checkmark & \checkmark & 74.5 & 92.6 & 65.8 & 46.6 & 71.2 & 33.4 \\
\bottomrule
\end{tabular}

\vspace{-4mm}
\label{tab:ablation}
\end{table*}
\medskip\noindent\textbf{Datasets}. 
UCF-101 \cite{soomro2012ucf101} contains 13,320 videos in 101 action categories and covers a wide range of scenes and recording conditions. 
HMDB-51 \cite{kuehne2011hmdb} includes 6,766 videos across 51 action categories and is considered challenging due to heavy camera motion and background variations. 
Something-Something-V2 (SS-V2) \cite{goyal2017something} is a larger dataset with 220,847 videos in 174 categories and focuses on fine-grained object interactions, requiring stronger temporal reasoning. 

Following the standard GCD protocol, we split each dataset into disjoint sets of known and novel classes. 
Specifically, UCF-101 is divided into 50 known and 51 novel classes, HMDB-51 into 25 known and 26 novel classes, and SS-V2 into 87 known and 87 novel classes. 
For each dataset, half of the videos from the known classes are treated as labeled, while the remaining videos from the known and all videos from the novel classes are used as unlabeled data.

\medskip\noindent\textbf{Evaluation Metrics}.
We follow standard practice \cite{vaze2022generalized, wen2023parametric} and report clustering accuracy (ACC) to measure model performance. 
Given the ground-truth labels $y^*$ and predicted labels $\hat{y}$, ACC is computed as
$
\text{ACC} = \frac{1}{P} \sum_{i=1}^{P} \mathbf{1}\left( y^*_i = p(\hat{y}_i) \right),
$
where $P = |\mathcal{D}^u|$ is the number of unlabeled samples and $p(\cdot)$ is the optimal permutation that aligns the predicted clusters with the ground-truth classes. 
Following prior work, we report ACC on three subsets: 
\textit{All} (all classes), \textit{Old} (only known classes), and \textit{New} (only novel classes).

\medskip\noindent\textbf{Implementation details.}
We adopt VideoMAE \cite{tong2022videomae} as the video backbone, initialized with Kinetics-400 \cite{kay2017kinetics} pretrained weights. MoE modules are inserted into the last $M = 2$ transformer blocks with $E = 8$ experts, and LoRA with rank $r = 4$ is applied to the $Q$, $K$, and $V$ projection matrices of all transformer blocks. Models are trained for 500 epochs with a batch size of 96 using the AdamW optimizer and a cosine learning rate schedule starting at $5 \times 10^{-4}$. We set $\beta = 0.05$ and $\lambda = 0.65$ for the auxiliary losses, and follow the official VideoMAE and SimGCD implementations for other details. All experiments are conducted on NVIDIA A6000 GPUs.

\vspace{-1mm}
\medskip\noindent\textbf{Baselines}. 
We compare VideoGCD against a range of state-of-the-art GCD methods, including ORCA \cite{cao2022openworld}, RankStats+ \cite{han2021autonovel}, GCD \cite{vaze2022generalized}, SimGCD \cite{wen2023parametric}, AdaptGCD \cite{qu2024adaptgcd}, and SPTNet \cite{wang2024sptnet}. We also compare with simple k-means \cite{arthur2006k} clustering on pretrained model's features.
All baselines are trained and evaluated under the same dataset splits and evaluation protocol described above for a fair comparison. 
More details are in the supplementary material.
\subsection{Main Results}
\medskip\noindent\textbf{Comparison with State-of-the-Art}. As shown in Table \ref{main_results}, VideoGCD consistently outperforms existing GCD methods across all three datasets. Specifically, on UCF-101, our approach achieves a substantial improvement of 10.4\% in overall accuracy compared to the next best method (SimGCD), and a 12.5\% gain on novel classes, highlighting the effectiveness of our \textit{novel-class balancing loss} to maintain strong learning path for unlabeled novel samples. 

Additionally, substantial improvements are observed on the challenging SS-V2 dataset, where VideoGCD improves overall accuracy by 3.6\% and novel-class accuracy by 4.5\% over SPTNet. Given the heavy camera motion and temporal variations in the SS-V2 dataset, this improvement shows the efficacy of the temporal OT loss in reducing prediction noise and enhancing temporal consistency. Performance on HMDB-51 shows similar gains where we outperform the second-best GCD method by 2.2\% in overall accuracy. These results show that the combination of MoE-based expert balancing, temporal consistency, and LoRA fine-tuning enables VideoGCD to effectively address the unique challenges of video GCD, particularly for novel classes where other methods struggle.

\begin{table}[ht]
\centering
\renewcommand{\arraystretch}{1.2}
\setlength{\tabcolsep}{4.3pt}
\caption{Effect of backbone adaptation across GCD methods}
\vspace{-1mm}
\resizebox{0.85 \linewidth}{!}{\begin{tabular}{
    l
    >{\columncolor{softblue}}c c c
    >{\columncolor{softblue}}c c c
}
\toprule
\textbf{Method} & \multicolumn{3}{c}{\textbf{UCF-101}} & \multicolumn{3}{c}{\textbf{HMDB-51}} \\
\cmidrule(lr){2-4} \cmidrule(lr){5-7}
 & All & Old & New & All & Old & New \\
\midrule

GCD & 54.7 & 61.7 & 51.4 & 32.7 & 41.3 & 28.1 \\

SimGCD & 69.9 & 88.2 & 61.2 & 43.3 & 66.1 & 30.9 \\
AdaptGCD & 71.3 & 88.7 & 63.0 & 44.7 & 66.4 & 33.0 \\
\rowcolor{softblue}
\textbf{VideoGCD} & \textbf{74.5} & \textbf{92.6} & \textbf{65.8} & \textbf{46.6} & \textbf{71.2} & \textbf{33.4} \\
\bottomrule
\end{tabular}}
\vspace{-4mm}
\label{tab:lora}
\end{table}

\begin{table}[ht]
\centering
\renewcommand{\arraystretch}{1.2}
\setlength{\tabcolsep}{3pt}
\caption{Comparison of different methods on UCF-101 and HMDB-51 when the number of classes is not known a priori.}
\vspace{-3mm}
\resizebox{0.85 \linewidth}{!}{\begin{tabular}{
    l c
    >{\columncolor{softblue}}c c c
    >{\columncolor{softblue}}c c c
}
\toprule
\textbf{Methods} & \textbf{Known $K$} 
& \multicolumn{3}{c}{\textbf{UCF-101}} 
& \multicolumn{3}{c}{\textbf{HMDB-51}} \\
\cmidrule(lr){3-5} \cmidrule(lr){6-8}
& & All & Old & New & All & Old & New \\
\midrule
GCD & \checkmark & 41.6 & 51.7 & 36.8 & 26.3 & 32.3 & 23.0 \\
SimGCD & \checkmark & \underline{64.1} & \underline{87.9} & 52.7 & 42.6 & \underline{70.6} & 27.4 \\
SPTNet & \checkmark & 61.4 & 87.7 & 48.8 & \underline{44.4} & 67.8 & \underline{31.8} \\
\rowcolor{softblue}
VideoGCD & \checkmark & \textbf{74.5} & \textbf{92.6 }& \textbf{65.8} & \textbf{46.6} & \textbf{71.2} & \textbf{33.4} \\
\midrule
GCD & -- (w/ Est.) & 39.0 &43.9 & 36.7 & 24.5 & 30.6 & 21.2 \\
SimGCD & -- (w/ Est.) & \underline{59.9} & 82.3 & \underline{49.2} & \underline{39.9} & \underline{63.9} & 26.9 \\
SPTNet & -- (w/ Est.) & 59.0 & \underline{87.5} & 45.4 & 34.5 & 39.0 & \underline{33.0} \\
\rowcolor{softblue}
VideoGCD & -- (w/ Est.) & \textbf{72.5} & \textbf{94.3} & \textbf{62.1} & \textbf{41.0} & \textbf{66.6} & \underline{27.2} \\

\bottomrule
\end{tabular}}
\label{tab:unknown_class}
\end{table}

\medskip\noindent\textbf{Ablation Study}.
In Table \ref{tab:ablation}, we conduct ablation experiments on UCF-101 and HMDB-51 to evaluate the contribution of each component in VideoGCD. Starting from the SimGCD baseline (row 1), adding LoRA fine-tuning (row 2) improves the accuracy on UCF-101 by 5.8\%, showing its importance in bridging the domain gap between pretraining and target datasets. 
Incorporating the MoE modules with the balancing loss $\mathcal{L}_{bal}$ (row 3) and the novel-class balancing loss $\mathcal{L}_{ncb}$ (row 4) further boosts performance, as these losses reduce expert collapse and ensure sufficient representation learning for novel classes. 
Finally, adding the temporal OT loss $\mathcal{L}_{tot}$ (row 5) provides the largest gain on HMDB-51 and additional improvements on UCF-101 by enhancing temporal consistency. 
The results confirm that each component contributes to GCD performance, with the full VideoGCD model achieving the best accuracy across all splits.

\medskip\noindent\textbf{Effect of Backbone Adaptation}.
Table~\ref{tab:lora} evaluates the impact of applying parameter-efficient backbone adaptation using LoRA to several GCD methods that typically update the final block of the network. We observe consistent improvements across all baselines when backbone adaptation is enabled. Importantly, even under the same backbone adaptation setting, our proposed VideoGCD framework continues to show substantial improvements over the baselines on both UCF-101 and HMDB-51. This indicates that the gains are not attributed to backbone adaptation alone, but stem from the proposed mechanisms for balanced expert specialization and temporal consistency.

\medskip\noindent\textbf{Robustness to Unknown Class Number}. 
In many real-world scenarios, the total number of categories is not known in advance. 
To evaluate this setting, we follow the class number estimation strategy used in ~\cite{vaze2022generalized} and compare VideoGCD with other methods when the number of classes is unknown. 
As shown in Table~\ref{tab:unknown_class}, the performance of all methods drops slightly compared to the setting with a known class number. 
However, VideoGCD still achieves the highest accuracy on both UCF-101 and HMDB-51, with 72.5\% and 41.0\% overall accuracy respectively, compared to the next best method SimGCD at 59.9\% and 39.9\%. 
This margin demonstrates that our approach remains consistently stronger even under this more challenging setting, confirming its robustness to the realistic case where the number of categories must be estimated rather than provided a priori.

\begin{table}[t]
\centering
\footnotesize
\setlength{\tabcolsep}{2pt}
\caption{Hyperparameter analysis on HMDB-51: Accuracy (\%) and improvement over the corresponding baseline setting on All, New, and Old classes.}
\vspace{-1mm}
\resizebox{0.85 \linewidth}{!}{\begin{tabular}{
    c c|
    >{\columncolor{softblue}}c c c|
    >{\columncolor{softblue}}c c c
}
\toprule
\multirow{2}{*}{\textbf{Hyperparam.}} & \multirow{2}{*}{\textbf{Value}} 
& \multicolumn{3}{c|}{\textbf{Accuracy} (\%)} 
& \multicolumn{3}{c}{\textbf{Improvement} (\%)} \\
 &  & All & New & Old & $\Delta$All & $\Delta$New & $\Delta$Old \\
\midrule
\multirow{5}{*}{$\beta$}
 & 0     & 43.3 & 30.9 & 66.1 & – & – & – \\
 & 0.01  & 47.1 & 32.0 & \textbf{75.0} & +3.8 & +1.1 & \textbf{+8.9} \\
 & 0.05  & 46.6 & 33.4 & 71.2 & +3.3 & +2.5 & +5.1 \\
 & 0.1   & \textbf{47.4} & \textbf{33.6} & 73.1 & \textbf{+4.1} & \textbf{+2.7} & +7.0 \\
 & 0.5   & 46.3 & 32.0 & 73.0 & +3.0 & +1.1 & +6.9 \\
\midrule
\multirow{5}{*}{$M$}
 & 0  & 43.3 & 30.9 & 66.1 & – & – & – \\
 & 1  & \textbf{46.6} & 32.2 & 73.4 & \textbf{+3.3} & +1.3 & \textbf{+7.3} \\
 & 2  & \textbf{46.6} & \textbf{33.4} & 71.2 & \textbf{+3.3} & \textbf{+2.5} & +5.1 \\
 & 4  & 46.1 & 31.1 & \textbf{73.7} & +2.8 & +0.2 & \textbf{+7.6} \\
 & 8  & 44.9 & 31.0 & 70.4 & +1.6 & +0.1 & +4.3 \\
\midrule
\multirow{5}{*}{$E$}
 & 0  & 43.3 & 30.9 & 66.1 & – & – & – \\
 & 2  & 46.1 & 31.9 & 72.2 & +2.8 & +1.0 & +6.1 \\
 & 4  & \textbf{47.1} & 32.9 & \textbf{73.3} & \textbf{+3.8} & +2.0 & \textbf{+7.2} \\
 & 6  & 46.3 & 32.0 & 73.0 & +3.0 & +1.1 & +6.9 \\
 & 8  & 46.6 & \textbf{33.4} & 71.2 & +3.3 & \textbf{+2.5} & +5.1 \\
\bottomrule
\end{tabular}}
\label{tab:hyperparam_ablation_combined}
\vspace{-5mm}
\end{table}
\medskip\noindent\textbf{Sensitivity to Hyperparameters.} 
We analyze the effect of three key hyperparameters, namely balancing loss weight $\beta$, number of MoE layers $M$, and number of experts $E$ on HMDB-51. Table~\ref{tab:hyperparam_ablation_combined} summarizes the results. In all cases, the model shows consistent gains over the baseline configuration, indicating robustness to a range of settings. Varying $\beta$ controls the relative strength of the regularization terms. Performance improves with small values and peaks at $\beta{=}0.1$, beyond which we observe marginal drops. This suggests that the regularization is beneficial but may lead to diminishing returns when overemphasized. For the number of MoE layers, $M{=}1$ and $M{=}2$ provide the best trade-off between complexity and performance. Deeper configurations ($M{=}4$, $M{=}8$) yield slightly lower results, but the degradation is minimal. Increasing the number of experts $E$ leads to consistent improvements up to $E{=}4$, after which the trend flattens or slightly declines. This behavior is likely due to over-partitioning or reduced expert utility at larger values. Overall, we observe that the proposed method is not overly sensitive to hyperparameter choices, and strong performance can be achieved with moderate settings.

\begin{table}[ht]
\centering
\renewcommand{\arraystretch}{1.1}
\setlength{\tabcolsep}{3.3pt}
\caption{\textbf{Effect of different temporal consistency losses.} Our partition-aware OT loss outperforms alternatives.}
\vspace{-1mm}
\resizebox{0.85 \linewidth}{!}{\begin{tabular}{
    l
    >{\columncolor{softblue}}c c c
    >{\columncolor{softblue}}c c c
}
\toprule
\textbf{Temporal Loss} & \multicolumn{3}{c}{\textbf{UCF-101}} & \multicolumn{3}{c}{\textbf{HMDB-51}} \\
\cmidrule(lr){2-4} \cmidrule(lr){5-7}
 & All & Old & New & All & Old & New \\
\midrule
No temporal loss & 72.7 & 91.0 & 64.0 & 45.5 & 71.7 & 31.8 \\
KL divergence & 73.7 & 91.7 & 65.3 & 45.9 & \textbf{71.8} & 32.5 \\
OT (uniform cost) & 73.9 & 92.1 & 65.1 & 46.2 & 70.6 & 33.2 \\
\rowcolor{softblue}
\textbf{OT (partition-aware)} & \textbf{74.5} & \textbf{92.6} & \textbf{65.8} & \textbf{46.6} & 71.2 & \textbf{33.4} \\
\bottomrule
\end{tabular}}

\vspace{-3mm}
\label{tab:tot_ablation}
\end{table}
\vspace{-1mm}
\medskip\noindent\textbf{Different Temporal Consistency Losses.} 
To assess the impact of our temporal consistency formulation, we compare the proposed partition-aware optimal transport (OT) loss against several alternatives in Table \ref{tab:tot_ablation}. Removing the temporal loss leads to a notable performance drop on both UCF-101 and HMDB-51, highlighting the importance of aligning frame-level routing decisions. Incorporating a simple frame-level KL divergence improves performance but remains suboptimal. Using OT with a uniform cost matrix yields further gains by allowing flexible alignment between routing distributions; however, it treats all expert transitions equally. In contrast, our partition-aware OT loss introduces a structured cost that penalizes transitions across partitions (e.g., from known to novel experts or vice versa) more severely. This design encourages temporally consistent and semantically coherent expert routing, resulting in performance improvements across nearly all subsets and datasets. These findings validate the effectiveness of our proposed loss formulation.

\begin{figure}
    \centering
    \includegraphics[width=1\columnwidth, trim={1.cm .5cm .5cm 1cm}]{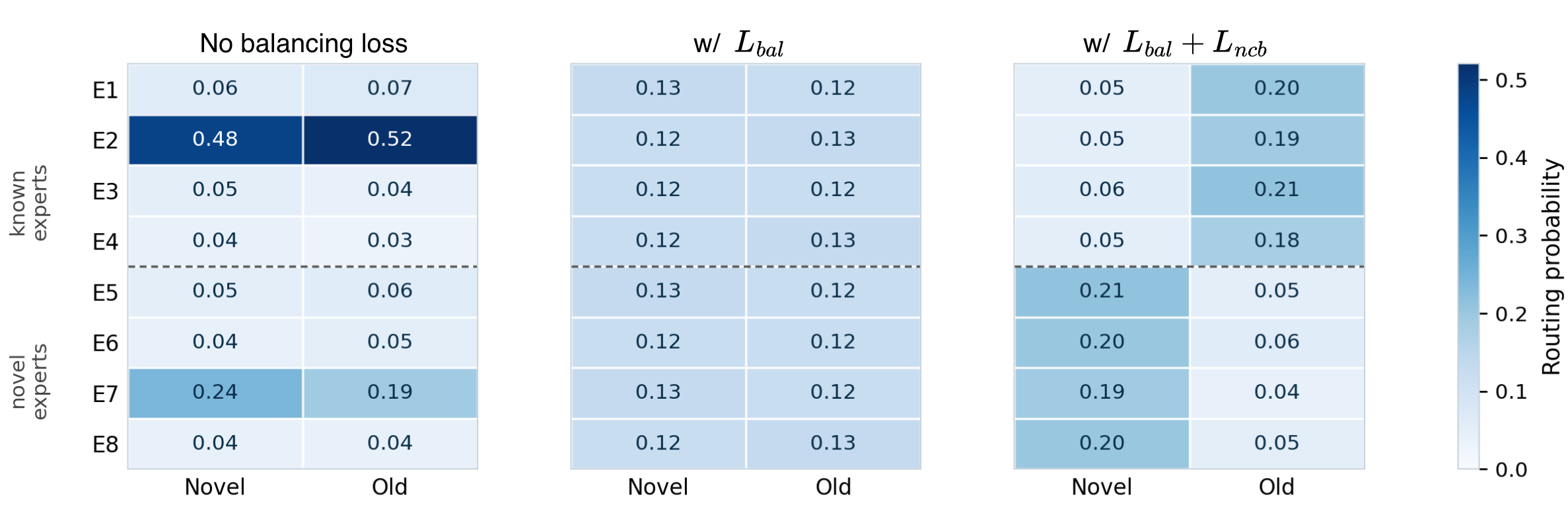}
    \vspace{-15pt}
    \caption{Empirical validation of balancing losses}
    \label{fig:balance}
\end{figure}

\medskip\noindent\textbf{Empirical Validation of Balancing Losses.} To validate the effect of our balancing losses, we visualize 
the per-expert routing probabilities for old and novel class samples on 
HMDB-51 in Fig.~\ref{fig:balance}. Without any balancing loss, routing 
collapses onto experts E2 and E7, leaving most experts 
unused. Adding $\mathcal{L}_\text{bal}$ resolves this collapse, 
producing a uniform distribution across all experts. 
Adding $\mathcal{L}_\text{ncb}$ further introduces a clear separation: 
Known experts (E1-E4) are activated for old samples, while novel experts (E5-E8) are activated for novel samples. These results confirm that $\mathcal{L}_\text{bal}$ effectively 
prevents expert collapse, and $\mathcal{L}_\text{ncb}$ ensures that 
novel-class samples are routed through dedicated experts, providing a 
reliable learning path for novel category discovery.

\begin{figure}
    \centering
    \includegraphics[width=0.8\columnwidth, trim={1.cm .5cm .5cm 1cm}]{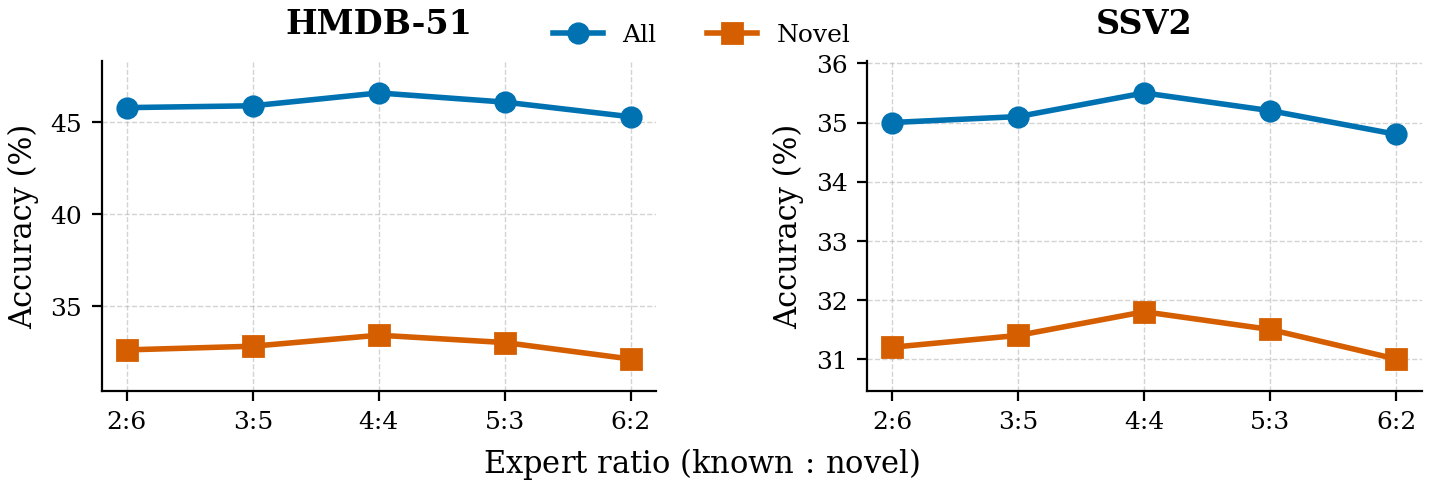}
    \vspace{-5pt}
    \caption{Known/Novel expert ratio in MoE}
    \vspace{-20pt}
    \label{fig:emperical}
\end{figure}

\medskip\noindent\textbf{Known/Novel Expert Ratio in MoE.}
In Fig.~\ref{fig:emperical}, we show that the ratio of known and novel experts in our MoE module is flexible and has little correlation with the known/novel class split in the GCD setting. Specifically, we vary the known-to-novel expert ratio from 1:3 to 3:1 on HMDB-51 and SSv2. We observe that the performance remains robust across all expert ratios and consistently outperforms the baseline without MoE.
\section{Conclusion}
In this paper, we introduced \textbf{VideoGCD}, a framework for generalized category discovery in video action recognition. Our approach is motivated by two key challenges in video GCD: the bias toward labeled known classes caused by stronger supervision, and the temporal instability of frame-level representations in video data. To address these issues, VideoGCD encourages balanced expert specialization while enforcing temporal consistency in the learning process, enabling the model to better allocate capacity for both known and novel categories. Comprehensive experiments on three benchmark datasets demonstrate that VideoGCD consistently improves performance over existing approaches and remains effective even when the number of categories is unknown. As part of future work, we plan to investigate the use of additional modalities such as optical flow and audio, as well as synthetic video data, to further enhance representation learning and advance generalized category discovery in videos.
{
    \small
    \bibliographystyle{ieeenat_fullname}
    \bibliography{main}
}

\end{document}